%% file: main.tex
\documentclass[conference]{IEEEtran}
\IEEEoverridecommandlockouts
\usepackage{cite}
\usepackage{amsmath,amssymb,amsfonts}
\usepackage{siunitx}
\usepackage{graphicx}
\usepackage{textcomp}
\usepackage{xcolor}
\usepackage{algorithm} 
\usepackage[hidelinks]{hyperref}
\usepackage{algpseudocode} 
\usepackage{booktabs, adjustbox}
\usepackage{enumitem}

\def\BibTeX{{\rm B\kern-.05em{\sc i\kern-.025em b}\kern-.08em
    T\kern-.1667em\lower.7ex\hbox{E}\kern-.125emX}}
\begin{document}

\title{\Large \bf Local\_INN: Implicit Map Representation and Localization with \\Invertible Neural Networks\\
{}
\thanks{}
}

\author{Zirui Zang, Hongrui Zheng, Johannes Betz, Rahul Mangharam \vspace{-2.5\baselineskip}
\thanks{All authors are with the University of Pennsylvania, Department of Electrical and Systems Engineering, 19104, Philadelphia, PA, USA. Emails: \{\tt\footnotesize zzang, hongruiz, joebetz, rahulm\}@seas.upenn.edu}
  
}

\maketitle

\input{abstract}
\input{intro.tex}
\input{related.tex}
\input{method.tex}
\input{exp.tex}
\input{discussion.tex}

\bibliographystyle{IEEEtran}
\bibliography{IEEEabrv,main}

\end{document}

%% file: abstract.tex
\begin{abstract}

Robot localization is an inverse problem of finding a robot's pose using a map and sensor measurements. In recent years, Invertible Neural Networks (INNs) have successfully solved ambiguous inverse problems in various fields. This paper proposes a framework that solves the localization problem with INN. We design an INN that provides implicit map representation in the forward path and localization in the inverse path. By sampling the latent space in evaluation, Local\_INN outputs robot poses with covariance, which can be used to estimate the uncertainty. We show that the localization performance of Local\_INN is on par with current methods with much lower latency. We show detailed 2D and 3D map reconstruction from Local\_INN using poses exterior to the training set. We also provide a global localization algorithm using Local\_INN to tackle the kidnapping problem. 


\end{abstract}

%% file: intro.tex
\section{Introduction}

Robot localization is a problem of finding a robot’s pose using a map and sensor measurements, like LiDAR scans. Any moving robot needs to interact with the physical world correctly. However, finding injective mappings between measurements and poses is difficult because sensor measurements from multiple distant poses can be similar. 

To solve this ambiguity, the widely adopted method, Monte Carlo Localization (MCL)\cite{dellaert1999monte, thrun2002probabilistic} uses random hypothesis sampling and sensor measurement updates to infer the pose. Other common approaches are to use Bayesian filtering\cite{fox2003bayesian} or to find better-distinguishable global descriptors on the map\cite{dube2020segmap,sarlin2019coarse}. Recent developments in localization research usually propose better measurement models or feature extractors within these frameworks. On contrary, this paper proposes a new approach to frame the localization problem as an ambiguous inverse problem and solve it with an invertible neural network (INN). We claim that INN is naturally suitable for the localization problem with many benefits, as we will show in this paper.

Robot localization is an inverse problem, which is when we are given a set of observations and try to find the causal factors. Usually, it’s easier to calculate the expected observations if given the causal factors. In the context of LiDAR-based localization, the robot's pose in the environment causes the particular scan measurements. In addition, when given a map, we can easily simulate LiDAR scans from any pose on the map. 

\begin{figure}[t]
\centering
\includegraphics[width=0.98\columnwidth]{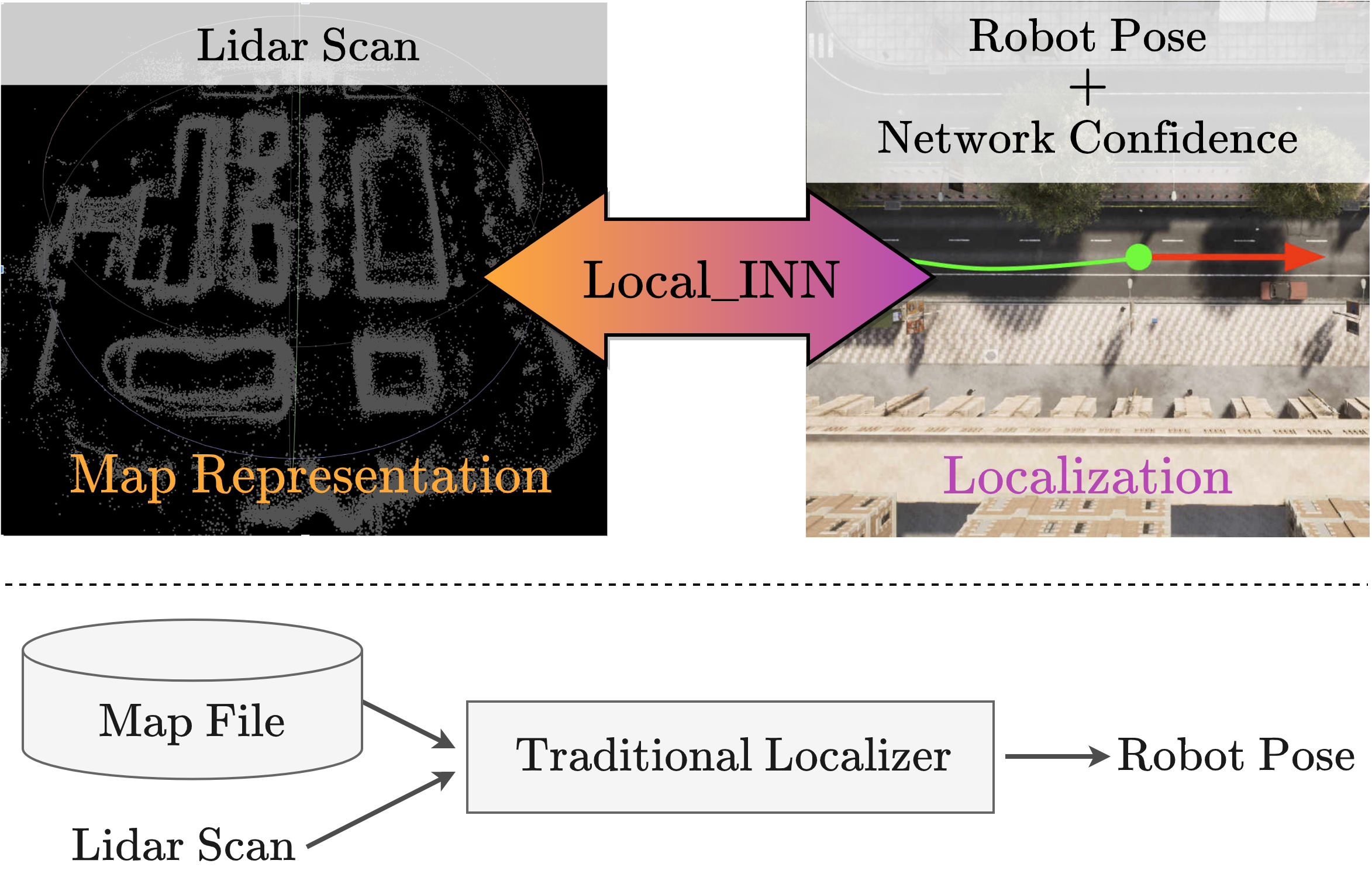}
\caption{Local\_INN is a framework of localization with invertible neural networks. Compared to current localization methods, Local\_INN stores map information within the neural network. Evaluation of Local\_INN in forward direction gives compressed map information, and in the reverse direction gives accurate localization with fast runtime and uncertainty estimation.}
\label{fig_local_inn}
\end{figure}

Invertible neural networks such as normalizing flows\cite{tabak2010density,dinh2016density,kingma2018glow,papamakarios2021normalizing} have been used to solve inverse problems in various fields\cite{ardizzone2018analyzing,ardizzone2019guided,adler2019uncertainty,xiao2020invertible,zhao2021invertible,WehRud2021}. It learns a bijective mapping between the source and target distributions with a series of invertible transformations. It uses a latent space to capture the lost ambiguous information during training. We use pose-scan data pairs to train such a bijective mapping. The forward path is from pose to scan and the reverse path is from scan to pose. Because INNs require the same input and output dimensions, we use a Variational Autoencoder (VAE)\cite{kingma2013auto} to reduce the dimension of the lidar scans and use Positional Encoding\cite{vaswani2017attention} to augment the dimension of the poses. With the help of conditional inputs, we can reduce the ambiguity of the inverse problem. In our case, we use zones in the map calculated from the previous pose of the robot as conditional input into the INN. During the evaluation, we sample the latent space to find the full posterior distribution of the pose, given a sensor measurement. We validated our method in localization experiments with 2D and 3D LiDARs, both in simulation and with real data. To summarize, this paper has four major contributions:

\begin{enumerate}[leftmargin=1.5em]
\item \textbf{Map Compression}: Local\_INN provides an implicit map representation and a localization method within one neural network. Map files are no longer needed when localizing.
    \item \textbf{Uncertainty Estimation}: Local\_INN outputs not just a pose but a distribution of inferred poses, the covariance of which can be used as the confidence of the neural network when fusing with other sensors, enhancing the overall robustness.
    \item \textbf{Fast and Accurate}: We demonstrate that the localization performance of Local\_INN is comparable to particle filter at slow speed and better at high speed with much lower latency with 2D LiDAR experiments.
    \item \textbf{Ability to Generalize}: We demonstrate that the framework of Local\_INN can learn complex 3D open-world environments and provides accurate localization. We also provide an algorithm for global localization with Local\_INN.
\end{enumerate}

%% file: related.tex
\section{Related Work}
Local\_INN sits at the intersection of two research fields: Localization and Normalizing Flows. In this section, we will briefly introduce both fields. 

\subsection{Lidar-based Localization}
Monte Carlo Localization (MCL)\cite{dellaert1999monte}, ever since its introduction, has been a popular localization framework for its reliable performance and the modularity to swap the motion or measurement model with any desired method. Many developments in localization seek to improve within the framework of MCL.\cite{zhang2018robust,chen2021icra,chen2020learning} Outside the framework of MCL, people have used methods such as Bayesian inference\cite{barsan2020learning}, RNNs\cite{lu2019l3,clark2017vidloc}, global descriptors\cite{uy2018pointnetvlad,cho2022openstreetmap}, or combining them\cite{sun2020localising}. We propose Local\_INN as a new framework of solve the problem. 

For learning-based localization methods, uncertainty estimations of the neural networks become a challenge.
There are efforts to approximate the uncertainty\cite{cai2019hybrid,kendall2016modelling,deng2022deep}, but it hasn't been widely applied. Local\_INN comes naturally with an uncertainty estimation due to the use of normalizing flows. 

Large map size is also becoming a burden as pointed out by \cite{wei2019learning}. After the advent of NeRF\cite{mildenhall2021nerf}, it was clear that neural networks are very capable of implicitly representing spatial information. There are developments in using neural networks for implicit map representation in the SLAM pipeline\cite{sucar2021imap,zhu2022nice}. Local\_INN builds on that while providing a method of localization.

\subsection{Normalizing Flows}

A normalizing flow is a series of invertible transformations that gradually transform a source data distribution into a target data distribution. Methods of achieving such bijective mappings have been developing rapidly in recent years\cite{papamakarios2021normalizing}. Real-valued non-volume preserving (RealNVP) transformations introduced by Dinh et al.\cite{dinh2016density} use coupling layers that are efficient to compute in both forward and reverse processes. Although newer normalizing flows have better expressiveness\cite{durkan2019neural,wu2020stochastic,zhang2021diffusion}, we choose to use RealNVP for its efficiency.

The framework of solving ambiguous inverse problems using normalizing flows was introduced by Ardizonne et al.\cite{ardizzone2018analyzing} and was later extended by \cite{ardizzone2019guided,winkler2019learning} to include a conditional input that is concatenated to the vectors inside the coupling layers. They proposed to use a latent variable to encode the lost information in training due to the ambiguity of the problem. During the evaluation, repeatedly sampling the latent variable can give the full posterior distribution given the input. In this paper, we added a VAE to the framework so that we can use high-dimensional input. The use of latent variables gives us distributions of estimated poses, which we can use to calculate the covariance.

%% file: method.tex
\section{Methodology}
LiDARs are widely used in moving robots and autonomous vehicles. 2D or 3D LiDARs produce one or multiple arrays of range distances with each value in the array being the distance from the robot to the closest obstacle at a certain angle. The localization problem with LiDAR is: given a LiDAR scan, find the robot's $[x, y]$ coordinates on the map and its heading $\theta$ relative to the $x$ axis of the map.
\begin{figure*}[t]
\centering
\includegraphics[width=1.94\columnwidth]{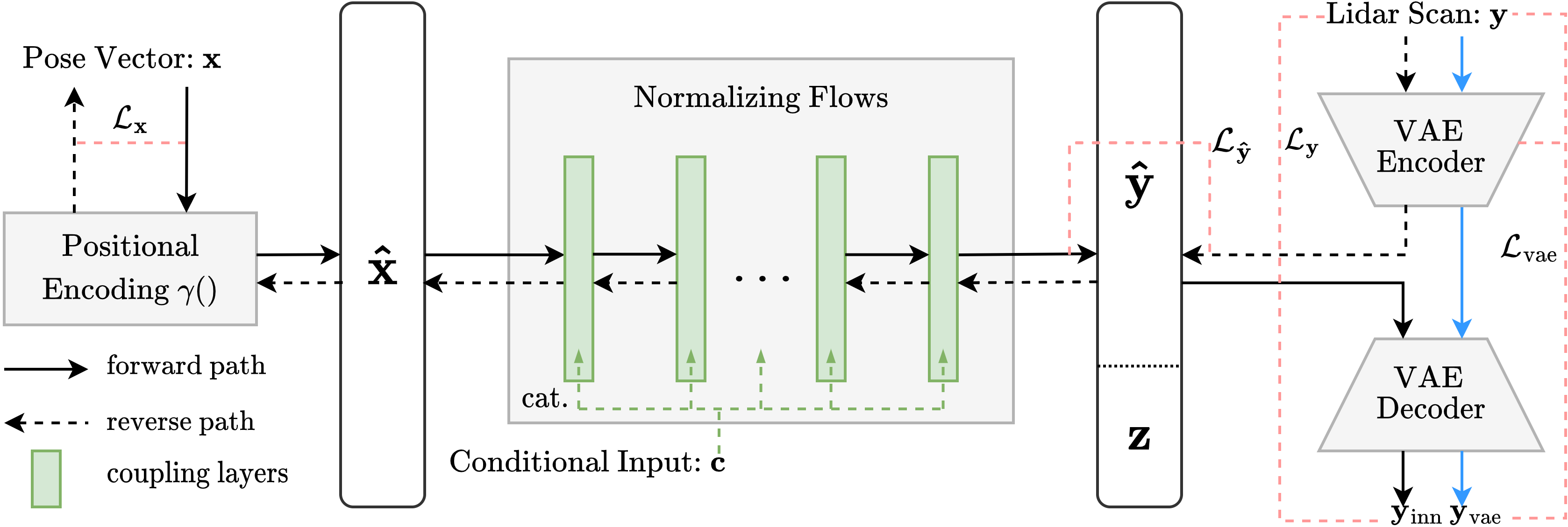}
\caption{Network Structure of the Local\_INN. The forward path (solid arrows) is from pose to LiDAR scan. The reverse path (dashed arrows) is from LiDAR scan to robot pose. Conditional input is calculated from the robot's previous pose. The INN used in this paper has 6 coupling layers and the VAE encoder and decoder have 2 layers of MLPs for 2D LiDARs and plus 6 layers of 2D convolutions for 3D LiDARs.}
\label{fig_local_inn}
\end{figure*}

We use normalizing flow to find a bijective mapping between a robot's pose vector $\mathbf{x} \in \mathbb{R}^3$ on the map and LiDAR scan vector $\mathbf{y} \in \mathbb{R}^{\text{angle}}$ with a latent vector $\mathbf{z} \in \mathbb{R}^6$. The forward path of the localization problem is easy, so we can simulate an infinite amount of pose-scan data pairs for training by randomly sampling the state space. We use a rounded pose (as in equation \ref{cond1}) of the robot to produce the conditional input $\mathbf{c} \in \mathbb{R}^{3}$ in the INN. This rounded pose can be computed during testing by rounding the robot's previous pose. Because INN requires the same input and output dimension, we use positional encoding to augment the pose vector $\mathbf{x}$ to $\mathbf{\hat{x}} \in \mathbb{R}^{6L}$, where $L$ is the level of the sine-cosine positional encoding \cite{vaswani2017attention}. On the LiDAR scan side, we use a VAE\cite{kingma2013auto} to encode the LiDAR scan $\mathbf{y}$ to $\mathbf{\hat{y}} \in \mathbb{R}^{6L-6}$, which is concatenated with latent vector $\mathbf{z} \sim \mathcal{N}(0,\,1)$. We use the latent vector to catch the full posterior distribution of $\mathbf{x}$ conditioned on $\mathbf{c}$ given $\mathbf{y}$. This can later be used to sample the covariance of the inferred pose vector.

\subsection{Conditional Normalizing Flow}

Normalizing flows contain a series of invertible transformations. We use the affine coupling block architecture introduced in Real-NVP\cite{dinh2016density} and extended by \cite{ardizzone2019guided,winkler2019learning} to incorporate a conditional input. The forward path of a single coupling block is:
\begin{equation}
\label{equ_affine}
\begin{split}
    \mathbf{v}_1 &= \mathbf{u}_1 \odot \exp(s_2(\mathbf{u}_2, \mathbf{\hat{c}})) + t_2(\mathbf{u}_2, \mathbf{\hat{c}}), 
    \\
    \mathbf{v}_2 &= \mathbf{u}_2 \odot \exp(s_1(\mathbf{v}_1, \mathbf{\hat{c}})) + t_1(\mathbf{v}_1, \mathbf{\hat{c}}).
    \end{split}
\end{equation}
The input $\mathbf{u}$ is split into two halves $\mathbf{u}_1$ and $\mathbf{u}_2$, which undergo affine transformations with scale coefficient $s_i$ and translation coefficient $t_i$ for $i\in\{1, 2\}$. Here $\odot$ is element-wise multiplication. The outputs $\mathbf{v}_1$ and $\mathbf{v}_2$ are then concatenated together before exiting this coupling block. The exponential function here is to eliminate zero outputs, which ensures invertibility. In the reverse direction, given $\mathbf{v}_1$ and $\mathbf{v}_2$, this structure is easily invertible without any computational overheads. Therefore, $s_i$ and $t_i$ are not required to be invertible and can be learned with neural networks. Multiple coupling blocks are connected to increase the expressiveness of the normalizing flows. After each coupling block, there is a predefined random permutation to shuffle the variables so that the splitting of the input vector is different for each block. We followed \cite{WehRud2021} to use two layers of MLP with ReLU activation in each affine coupling block and used a parameterized soft clamping mechanism to prevent instabilities. Let's denote the forward and reverse path of the INN network with $h_{\text{inn}}^{\mathit{forward}}$, $h_{\text{inn}}^{\mathit{reverse}}$.

To deal with the inverse ambiguity due to map symmetry, a rounded pose computed from the robot's previous pose $\mathbf{x}^{\text{pre}}$ is passed through a positional encoding $\gamma(\cdot)$, then encoded by a separate MLP $h_{\text{cond}}$ before concatenating to $\mathbf{u}_i$ or $\mathbf{v}_i$ in the coupling block:
\begin{equation}
\mathbf{c} = \frac{\lceil N \mathbf{x^{\text{pre}}}\rfloor}{N}, 
\mathbf{\hat{c}}= h_{\text{cond}}(\gamma(\mathbf{c})).
\label{cond1}
\end{equation}
During training, $\mathbf{x}^{\text{pre}}$ is approximated by adding a zero mean Gaussian noise to the ground truth pose:
\begin{equation}
\mathbf{x}^{\text{pre}}_{\text{training}} = \mathbf{x} + \delta, \delta \sim \mathcal{N}(0,\,\sigma^2).
\label{cond2}
\end{equation}
 The rounded previous states essentially divide the state space into $N^3$ zones, and which zone the robot previously existed in is provided to the INN as conditional input. The Gaussian noise during training ensures that it's okay for the $\mathbf{x^{\text{pre}}}$ near zone boundaries to be rounded into either neighboring zones. Depending on the map, $\sigma^2$ for $[x, y, \theta]$ and integer parameter $N$ needs to be picked. We picked $\sigma^2$ around 0.5 meters and $N = 10$ for all our experiments, which means the zones are quite large for the 3D maps.

\subsection{Positional Encoding}

Positional encoding was used in \cite{vaswani2017attention} \cite{mildenhall2021nerf} to boost the performance of the neural network in fitting high-frequency information. The positional encoding $\gamma(\cdot)$ we used maps from $\mathbb{R}$ to $\mathbb{R}^{2L}$ with increasing frequencies:
\begin{equation}
\begin{split}
\gamma(p)= (&\sin(2^0\pi p), \cos(2^0\pi p), \dotsc,\\  &\sin(2^{L-1}\pi p), \cos(2^{L-1}\pi p)).
\end{split}
\end{equation}
When applied to pose vectors, function $\gamma(\cdot)$ is applied separately to $[x, y, \theta]$. Pose vector $\mathbf{x}$ is encoded with $L=10$ and the conditional input is encoded with $L=1$. All variables are normalized to $[0, 1)$ before being applied to $\gamma(\cdot)$. We observed that adding positional encoding directly helps the forward path by augmenting the 3-dimensional input, which in turn helps the reverse training as well. 


\subsection{Variational Autoencoder}

LiDARs produce hundreds to thousands of range data points per channel. Due to the input and output dimension requirement of the INN, putting everything into the INN would vastly increase the size of the network without proportional benefit. On the other hand, sub-sampling LiDAR scans increase susceptibility to noisy or invalid LiDAR points. Therefore, to fully utilize the LiDAR scans points and simultaneously limit the network size, we use a VAE to first encode the LiDAR scans into a multivariate Gaussian latent space with mean $\boldsymbol{\mu}_{\text{vae}}$ and variance $\boldsymbol{\sigma}^2_{\text{vae}}$.

The encoder $h_{\text{vae}}^{\mathit{encode}}$ of the VAE has one-layer MLP with ReLU that is connected to the input, and two separate one-layer MLPs for encoding $\boldsymbol{\mu}_{\text{vae}}$ and $\boldsymbol{\sigma}^2_{\text{vae}}$. Then, the encoder outputs by random sampling the encoded distribution:
\begin{equation}
\mathbf{\hat{y}} \sim \mathcal{N}(\boldsymbol{\mu}_{\text{vae}},\,\boldsymbol{\sigma}^2_{\text{vae}})
\end{equation}

The decoder $h_{\text{vae}}^{\mathit{decode}}$ of the VAE has two layers of MLP. The first one is with ReLU and the second is with Sigmoid.

\subsection{Optimization}

The guaranteed invertibility of INN means that we can do bi-directional training by optimizing loss from both sides of the network. We train both the forward and reverse paths with supervised losses. In each epoch, the forward and reverse paths are both calculated and gradients are added together before an optimizer step. 

The VAE network is responsible for encoding and reconstructing the LiDAR scans:
\begin{equation}
\begin{split}
\mathbf{\hat{y}}_{\text{vae}} &=h_{\text{vae}}^{\mathit{encode}}( \mathbf{y}_{\text{gt}}),   
\\
\mathbf{y}_{\text{vae}} &=h_{\text{vae}}^{\mathit{decode}}( \mathbf{\hat{y}}_{\text{vae}}),
\end{split}
\label{process_vae}
\end{equation}
which is optimized for the commonly used ELBO loss:
\begin{equation}
\mathcal{L}_{\text{vae}} = \| \mathbf{y}_{\text{gt}} - \mathbf{y}_{\text{vae}} \|_1 + \lambda_{\text{KL}} \text{KL}( \mathcal{N}(\boldsymbol{\mu}_{\text{vae}},  \boldsymbol{\sigma}^2_{\text{vae}}),  \mathcal{N}(0, 1)),
\label{loss_vae}
\end{equation}
where $\lambda_{\text{KL}}$ is a weight for the KL divergence term.
The VAE is trained together with the INN.

Each epoch of training the INN starts with evaluating the encoder of the VAE with ground truth LiDAR scans $\mathbf{y}_{\text{gt}}$ to get the encoded scans $\mathbf{\hat{y}}_{\text{vae}}$, and evaluating the decoder of the VAE with the output of the encoder, as in (\ref{process_vae}). $\mathcal{L}_{\text{vae}}$ is calculated as in (\ref{loss_vae}). The next step is to evaluate the forward path of the INN with $\mathbf{\hat{x}}_{\text{gt}}$ to get the forward output:
\begin{equation}
[\mathbf{\hat{y}}_{\text{inn}}, \mathbf{z}_{\text{inn}}] = h_{\text{inn}}^{\mathit{forward}}(\mathbf{\hat{x}}_{\text{gt}}, \mathbf{\hat{c}}).
\end{equation}
We then evaluate the decoder of VAE again with output from the INN forward path:
\begin{equation}
\mathbf{y}_{\text{inn}} = h_{\text{vae}}^{\mathit{decode}}(\mathbf{\hat{y}}_{\text{inn}}),
\end{equation}
and calculate a loss on the LiDAR scan output:
\begin{equation}
\mathcal{L}_{\mathbf{y}} = \|\mathbf{y}_{\text{gt}} - \mathbf{y}_{\text{inn}}\|_1.
\end{equation}
We also calculate a loss that matches the output of the INN forward path with the output of the VAE encoder:
\begin{equation}
\mathcal{L}_{\mathbf{\hat{y}}} = \|\mathbf{\hat{y}}_{\text{vae}} - \mathbf{\hat{y}}_{\text{inn}}\|_1.
\end{equation}

For the reverse path of the INN, we first evaluate with the encoded scan from VAE encoder concatenated by the latent vector generated by the forward path: [$\mathbf{\hat{y}}_{\text{vae}}$, $\mathbf{z}_{\text{inn}}$]. This produces a predicted pose we call $\mathbf{\hat{x}}_{\text{inn},\text{0}}$:
\begin{equation}
\mathbf{\hat{x}}_{\text{inn},\text{0}} = h_{\text{inn}}^{\mathit{reverse}}([\mathbf{\hat{y}}_{\text{vae}}, \mathbf{z}_{\text{inn}}], \mathbf{\hat{c}}).
\end{equation}
We calculated a L1 loss between $\mathbf{\hat{x}}_{\text{inn},\text{0}}$ and the ground truth:
\begin{equation}
\mathcal{L}_{\mathbf{\hat{x}},0} = \|\mathbf{\hat{x}}_{\text{gt}} - \mathbf{\hat{x}}_{\text{inn},\text{0}}\|_1.
\end{equation}
Following \cite{WehRud2021}, the intuition of this reverse evaluation is to link the encoded scans plus the predicted latent vector to the single corresponding pose in the ambiguous inverse problem. 

To capture the full posterior, we then sample $m$ latent vectors $\mathbf{z} \sim \mathcal{N}(0,\,1)$ and evaluate the reverse path using the sampled latent vectors combined with $\mathbf{\hat{y}}_{\text{vae}}$. 
\begin{equation}
\mathbf{\hat{x}}_{\text{inn},i} = h_{\text{inn}}^{\mathit{reverse}}([\mathbf{\hat{y}}_{\text{vae}}, \mathbf{z}_i], \mathbf{\hat{c}}),\text{for } i=1\dots m.
\end{equation}
This generates $m$ poses and we select the minimum of the L1 losses as the second part of the reverse loss:
\begin{equation}
\mathcal{L}_{\mathbf{\hat{x}},i} = \min_{i = 1 \dots m} \|\mathbf{\hat{x}}_{\text{gt}} - \mathbf{\hat{x}}_{\text{inn},i}\|_1.
\end{equation}

Overall, the training loss of the whole network is:
\begin{equation}
\mathcal{L}_{\text{all}} = \mathcal{L}_{\text{vae}}
+ \mathcal{L}_{\mathbf{y}}
+ \lambda_{\mathbf{\hat{y}}}\mathcal{L}_{\mathbf{\hat{y}}}
+ \mathcal{L}_{\mathbf{\hat{x}},0}
+ \mathcal{L}_{\mathbf{\hat{x}},i},
\end{equation}
where $\lambda_{\mathbf{\hat{y}}}$ is the weight for $\mathcal{L}_{\mathbf{\hat{y}}}$.

%% file: exp.tex
\section{Experiments}

\subsection{2D LiDAR Localization and real-world robot}

\begin{table*}[t]
\centering
\setlength{\tabcolsep}{2pt}
\caption{Map Reconstruction and Localization Errors with 2D LiDAR}
\begin{tabular}{ l c c c c c c}
\toprule 
\multicolumn{1}{c}{} & 
\multicolumn{2}{c}{Race Track (Simulation)} & 
\multicolumn{2}{c}{Hallway (Real)} & 
\multicolumn{2}{c}{Outdoor (Real)}  \\ 
\midrule

\begin{tabular}[c]{@{}c@{}}Original Map \\ \textcolor{orange}{Reconstruction} \\
\textcolor{violet}{Test Trajectory}
\end{tabular}
& \multicolumn{2}{c}{\adjincludegraphics[valign=c,width=0.48\columnwidth]{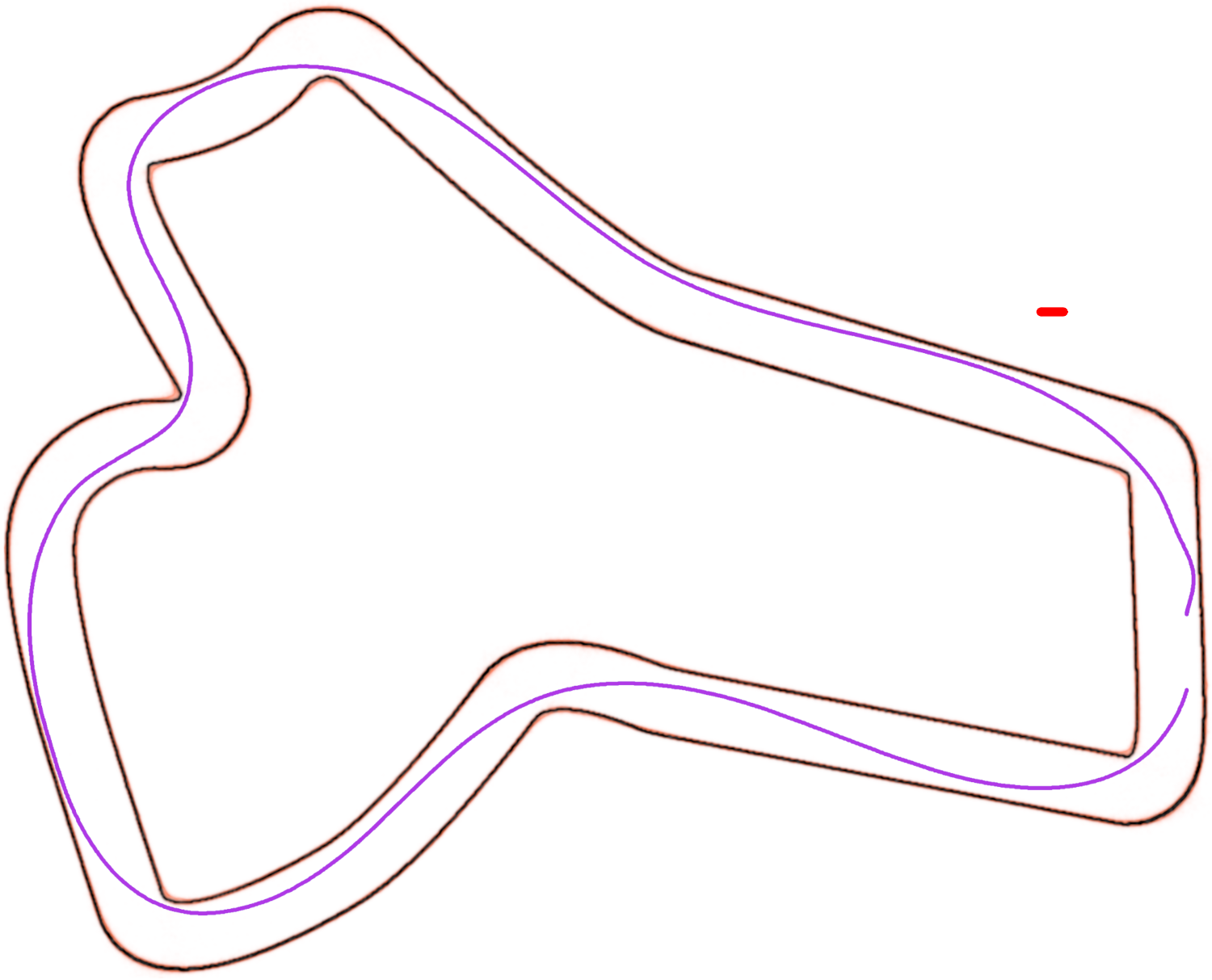} }
& \multicolumn{2}{c}{\adjincludegraphics[valign=c,width=0.48\columnwidth]{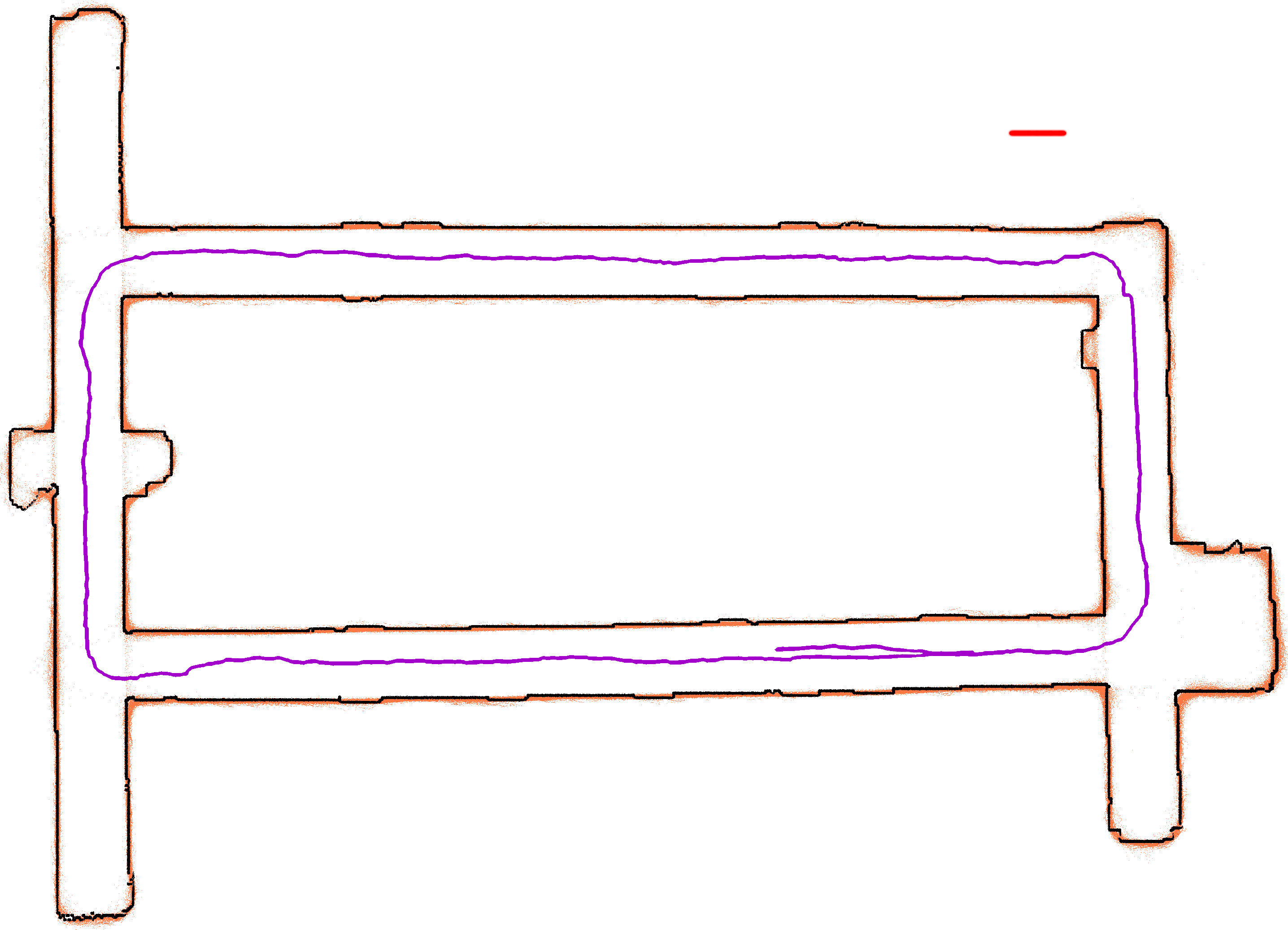} }
& \multicolumn{2}{c}{\adjincludegraphics[valign=c,width=0.48\columnwidth]{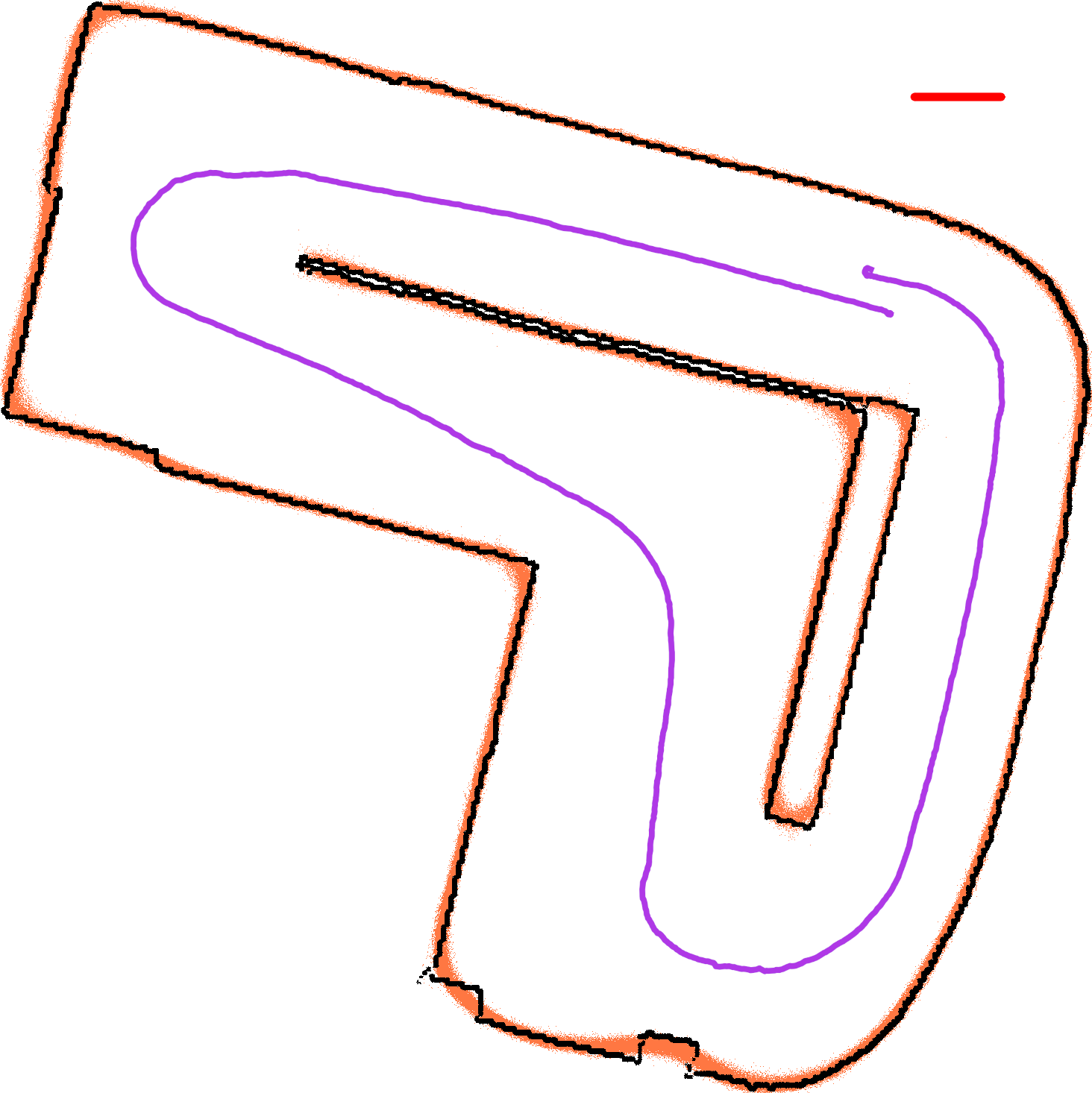} }
\\ \midrule
 
& $xy$(m)& $\theta$($^\circ$) 
& $xy$(m)& $\theta$($^\circ$) 
& $xy$(m)& $\theta$($^\circ$) 
\\ \midrule
Online PF (1m/s)
& $0.045 \pm \mathbf{0.058}$& $0.400 \pm 0.512$
& $\mathbf{0.039} \pm \mathbf{0.066}$& $\mathbf{0.482} \pm 0.808$ 
& $\mathbf{0.013} \pm \mathbf{0.018}$& $\mathbf{0.358} \pm \mathbf{0.456}$   \\

Local\_INN (1m/s)
& $0.050 \pm 0.102$& $0.201 \pm 0.532$  
& $0.196 \pm 0.433$& $0.528 \pm \mathbf{0.792}$
& $0.034 \pm 0.047$& $0.924 \pm 1.130$  \\

↑ + EKF
& $\mathbf{0.039} \pm 0.077$& $0.182 \pm 0.464$
& $0.093 \pm 0.139$& $0.536 \pm 0.797$
& $0.034 \pm 0.047$& $0.917 \pm 1.129$ \\

↑ + TensorRT
& $0.039 \pm 0.076$& $\mathbf{0.177} \pm \mathbf{0.443}$
& $0.104 \pm 0.159$& $0.547 \pm 0.802$
& $0.033 \pm 0.046$& $0.930 \pm 1.142$

\\ \midrule

Online PF (5m/s)
& $0.139 \pm 0.168$& $1.463 \pm 2.107$
& $\mathbf{0.071} \pm \mathbf{0.117}$& $0.943 \pm 1.738$
& $0.033 \pm 0.047$& $0.940 \pm 1.371$
\\
Local\_INN+EKF (5m/s)
& $\mathbf{0.034} \pm \mathbf{0.056}$& $\mathbf{0.133} \pm \mathbf{0.284}$
& $0.100 \pm 0.147$& $\mathbf{0.565} \pm \mathbf{0.900}$
& $\mathbf{0.032} \pm \mathbf{0.046}$ & $\mathbf{0.915} \pm \mathbf{1.130}$






\\ \bottomrule
\end{tabular}
\label{table_2d}
\end{table*}

We first validate the proposed method of localization with three different 2D LiDAR maps. The first map is a race track in simulation, and the second and third maps are real-world indoor hallway and outdoor environments mapped using the ROS SLAM toolbox with an \href{https://f1tenth.org/}{F1TENTH} racecar\cite{pmlr-v123-o-kelly20a}, which is a 1/10 scale autonomous racing car equipped with a Hokuyo 30LX LiDAR and a NVIDIA Jetson Xavier NX board. To collect training data, we uniformly sample $[x, y, \theta]$ on the drivable surface of each map, and use a 2D LiDAR simulator to find the corresponding LiDAR ranges. This means the trained network will be able to localize everywhere on the map. We collect 100k data pairs and train a separate network for each map.

To test the localization performance, we localize a car robot following a test trajectory in each environment and compare the inferred pose with the ground truth. For the real maps, we train with simulated data but test using real LiDAR data on the \href{https://f1tenth.org/}{F1TENTH} car driving in indoor and outdoor environments. We approximate the ground truth poses using a particle filter with the full LiDAR inputs and running it offline on a desktop with an infinite compute budget. For a baseline, we configured a GPU-accelerated particle filter \cite{walsh17}, so that it can run around the same frequency as the Local\_INN on the Jetson NX.

In these experiments, we use 270 points for each LiDAR scan $\mathbf{y}$ covering 270 degrees in front of the LiDAR, following the FoV of the Hokuyo LiDARs. $\mathbf{\hat{y}}$ is set to have 54 dimensions and $\mathbf{z}$ to have 6 dimensions. The encoder of the VAE has one layer of MLP before regressing the $\boldsymbol{\mu}_{\text{vae}}$ and $\boldsymbol{\sigma}^2_{\text{vae}}$ with separate MLP layers. The decoder has two layers of MLP converting 54-dimension $\mathbf{\hat{y}_{\text{vae}}}$ back to 270 ranges points. The INN network has 6 coupling layers, each having separate MLP layers for scale and translation coefficients. We trained the network with batchsize of 500 and with a learning rate that starts from $\num{1e-3}$ and exponentially decays to $\num{5e-5}$ in 600 epochs.

The map reconstruction is qualitatively evaluated by calculating the forward path with additionally random sampled test poses. The inferred LiDAR ranges are then converted into the map frame and accumulated to produce an occupancy map. The orange dots in table \ref{table_2d} are reconstructed maps. We can see the reconstructed map largely overlaps with the real map with some losses in high spatial-frequency information at hard corners. The red bar on the upper right corner of each map is an indicator for 1 meter.

During the inference of the reverse path, we sample latent vector $\mathbf{z}$ and calculate a batch of inferred poses. We can use the covariance of each batch as the confidence of the network. To demonstrate this, we use an Extended Kalman Filter to fuse the network outputs with vehicle odometry. The EKF uses a kinematic bicycle model as the motion model, and the pose output and covariance from the INN as the observation model.

Table \ref{table_2d} presents localization absolute mean and RMS errors in each environment. We see that not only the localization performance is comparable to particle filter, but the error and RMS also do not increase with vehicle speed. On the contrary, we see the error increase with the particle filter. This is because Local\_INN does not directly rely on the smoothness of the state's history, but only relies on the zoning provided by the previous state. Table \ref{table_2d_runtime} compares the runtime of the Local\_INN with the GPU-accelerated particle filter we used. We are comparing the latency of Local\_INN and particle filter. Other latencies are not accounted for. With runtime optimizations like TensorRT, Local\_INN can output localization results with much lower latency than particle filter with almost no decrease in performance, which is crucial in latency-sensitive applications such as high-speed racing \cite{betz_autonomous_2022}.

\begin{table}[h]
\vspace{-10pt}
\centering
\caption{Runtime Comparisons on NVIDIA Jetson NX}
\begin{tabular}{ l | c }
\toprule 
Online PF & $45$ Hz\\
Local\_INN (Pytorch)& $48$ Hz\\
Local\_INN+TensorRT & $\mathbf{270}$ Hz
\\ \bottomrule
\end{tabular}
\label{table_2d_runtime}
\vspace{-10pt}
\end{table}

\subsection{3D Open Space LiDAR Localization}

We then extended our experiments to using 3D LiDAR data, for which we also have three different environments: Town 10 in the CARLA simulator\cite{dosovitskiy2017carla}, KAIST in Mulran dataset\cite{kim2020mulran}, and Columbia Park in Apollo dataset\cite{lu2019l3}. For CARLA, we used the simulator to sample all drivable surfaces in the town. To fully train the Local\_INN, simulating a large amount of data from the map is preferred. But for comparison with existing works, we just used provided data points for Mulran and Apollo datasets. When testing the network, we report numbers from in-session and out-session localization. For in-session results, in CARLA, we have additionally sampled points; in Mulran and Apollo, we randomly picked and set aside 20\% of the dataset for testing. For out-session tests, the network is tested with sequences that are captured at another date. We provide in-session performances to show that the network is able to interpolate between the training data.

We treat 3D LiDAR scans as range images for the 3D experiments. To correctly reconstruct the out-of-range LiDAR points, we added a mask layer to the range images, and an L2 loss on it. The structure and dimension of the network are mostly unchanged for the 3D experiments. The only additions are 6 layers of 2D convolution and transpose convolution layers to the encoder and decoder of the VAE for the range images.


\begin{figure*}[t]
\centering
\includegraphics[width=1.8\columnwidth]{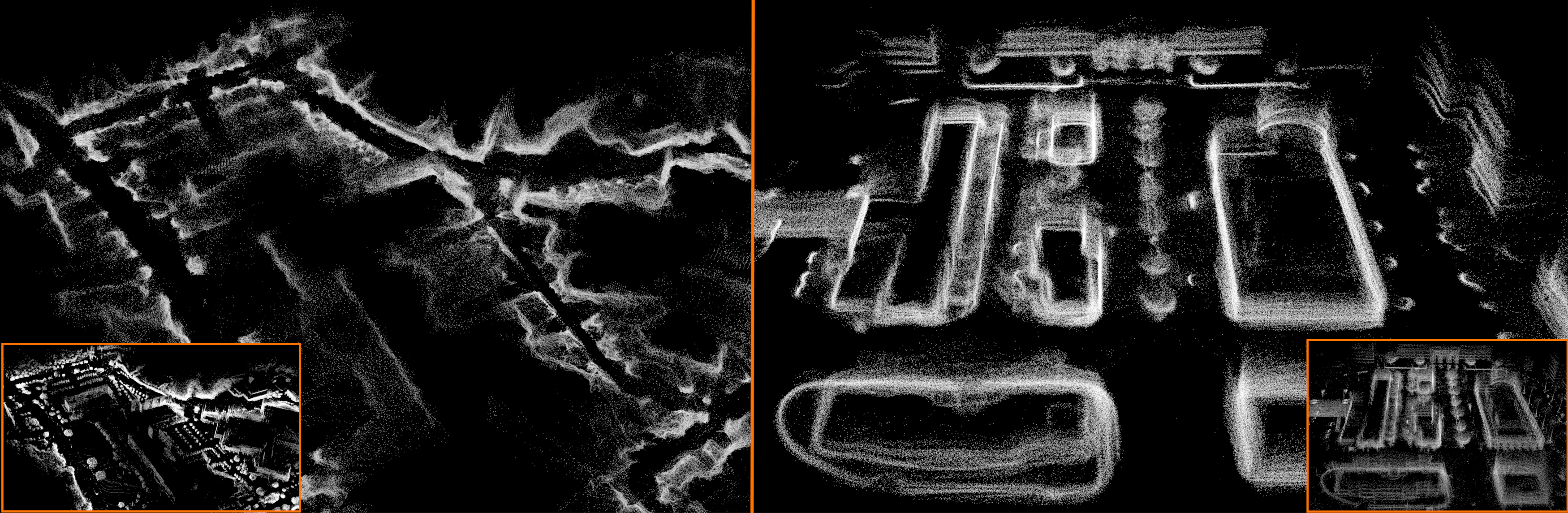}
\caption{3D Map Reconstruction for Mulran and CARLA. Orange boxed thumbnails are original maps. Reconstructions are produced by evaluating the forward path of the Local\_INN with poses exterior to the training set.}
\label{fig_3d_recon}
\end{figure*}

The quality of the map reconstruction is again qualitatively examined as some examples are shown in Fig. \ref{fig_3d_recon}. Because we simulated many more data points from the CARLA environment, we can see the reconstruction is very close to the original point cloud.

Table \ref{table_3d_error} shows a comparison of RMS errors between our method and existing works in localization experiments with 3D LiDARs. Due to the simplicity of our 3D setup, we are comparing to a method from Chen et al\cite{chen2021icra} that only uses range images from 3D LiDARs, and a method from Yin et al\cite{yin2021rall} that also uses a neural network with convolution layers to treat LiDAR information. We can see that our results on par with the state-of-the-art.

\begin{table}[H]
\centering
\caption{Comparison of Localization RMS Errors with 3D LiDAR}
\begin{tabular}{ l | c c c }
\toprule 
Methods ($xy$[m], $\theta$[$^\circ$]) & CARLA & Mulran & Apollo \\
\midrule 
Local\_INN in-session
& $0.27, 0.12$ 
& $0.29, 0.24$ 
& $0.50, 0.26$ \\
\midrule
Local\_INN out-session
& $-, -$
& $1.41, 1.00$ 
& $1.22, 0.53$  \\
Chen et al.
& $0.48, 3.87$ 
& $0.83, 3.14$ 
& $0.57, 3.40$  \\
RaLL (Yin et al.)
& $-, -$
& $1.27, 1.50$ 
& $-, -$ 
\\ \bottomrule
\end{tabular}
\label{table_3d_error}
\vspace{-5pt}
\end{table}

\subsection{Global Localization}

Global localization is needed when a robot starts with an unknown pose or when the robot encounters the kidnapping problem. MCL algorithms usually do global localization by spreading the covariance all around the map and using iterations of control inputs and measurements to decrease the covariance. For Local\_INN, the global localization process mainly involves simultaneously tracking multiple assumptions of zoning on the map and a selection process to narrow down the assumptions.

Algorithm \ref{alg_global} shows our global localization process. We track a set $\mathcal{C}$ of $n$ conditional inputs, each with a weight $w_i$ for $i = 1\dots n$. The set $\mathcal{C}$ is initialized by randomly sample $N$ states in the state space $\mathcal{S}$. We set the total number of latent vectors $\mathbf{z}$ sampled from normal distribution as $nM$ for a constant $M$. Initially, every $\mathbf{c}_i \in \mathcal{C}$ has the same weight $w_i$, so each one gets $M$ samples of $\mathbf{z}$. 

\begin{algorithm}[h]
\caption{Local\_INN Global Localization} 
\begin{algorithmic}[1]
\State $n \leftarrow N, m_i \leftarrow M, w_i \leftarrow 1/M$ for $i=1\dots n$
\State $\mathcal{X^{\text{rand}}} \leftarrow$ random\_sample($\mathcal{S}, n$)
\State $\mathcal{C}_0 \leftarrow$ convert\_to\_cond\_inputs($\mathcal{X^{\text{rand}}}$)


\While {\textit{new LiDAR scan $\mathbf{y}_{t+1}$ coming}}

\For{$i=1\dots n_{t}$}
\State $\mathbf{x}_{t+1,i} \leftarrow$ Local\_INN\_reverse($\mathbf{y}_{t+1}, \mathbf{c}_i, m_i$)

\State $\mathcal{X}_{t+1}\text{.append(}\mathbf{x}_{t+1,i}\text{)}$

\State $\mathbf{y}_{\text{inn},i} \leftarrow$ Local\_INN\_forward($\mathbf{x}_{t+1,i}, \mathbf{c}_i$)

\State $w_i \leftarrow 1/\|\mathbf{y}_{\text{inn},i} -  \mathbf{y}_{t+1}\|_1$ 

\EndFor

\State $\mathcal{C}_{t+1} \leftarrow$ convert\_to\_cond\_inputs($\mathcal{X}_{t+1}$)
\State $n_{t+1} \leftarrow |\mathcal{C}_{t+1}|$
\For{$i=1\dots n_{t+1}$}
\State $m_i \leftarrow  \text{normalized(} w_i\text{)}n_{t+1}M$
\EndFor

\EndWhile
\end{algorithmic} 
\label{alg_global}
\end{algorithm}


When a new LiDAR scan arrives, for each $\mathbf{c}_i \in \mathcal{C}$, we evaluate the reverse path of the Local\_INN with $m_i$ samples of latent vector $\mathbf{z}$. The output poses from Local\_INN become the next $\mathcal{C}$. We then update the weight $w_i$ for every $\mathbf{c}_i$ using the reciprocal of the scan error, calculated with the current sensor measurement, and inferred LiDAR scan from evaluating the forward path of the Local\_INN. We favor the $\mathbf{c}_i \in$ 
$\mathcal{C}$ that have higher weights by redistributing $\mathbf{z}$ samples based on the weights. Those with higher weights will have more $\mathbf{z}$ samples, which in turn may result in better pose estimations. It also should be noted that the size of $\mathcal{C}$ will decrease as iterations go because repeated elements in $\mathcal{C}$ are combined. Hence, we design a selection process to find the best-fit candidate. Lastly, we record the accumulated weights for every iteration and the $\mathbf{c}_i$ with the highest accumulated weights will be the most likely zone that the robot exists in.

We test out the above algorithm with different environments. We define a \texttt{Converged} as the correct pose having the highest weight and a \texttt{Tracking} as the correct pose within the top 5 on the tracking list. Table \ref{table_global} presents the percentage of \texttt{Converged} cases,  \texttt{Tracking} cases, and the absolute mean errors if the correct pose is picked or in tracking at the 10th iteration. The starting poses are randomly picked and the rates are out of 2k tests in each environment. We use the test trajectory for the 2D maps and out-of-session test sets for the 3D maps. The result shows the neural network can quickly identify correct poses with only 10 LiDAR scans. We can also see in the Hallway map, that the convergence of the assumptions is slower, which is expected in this highly symmetrical environment. As the algorithm keeps iterating with new LiDAR data, it will eventually converge to the correct pose.

\begin{table}[h]
\vspace{-8pt}
\centering
\caption{Global Localization Success Rates in Different Environments at Iteration 10}
\begin{tabular}{l | c c c}
\toprule 
Map & Converged & Tracking & $\Delta_{xy}$, $\Delta_{\theta}$ \\
\midrule 
Race Track & 79.5\% & 99.5\% & 0.075, 0.274\\
Hallway & 66.4\% & 91.1\% & 0.258, 0.538\\
Outdoor & 98.5\% & 100\% & 0.049, 0.911\\
Mulran & 93.5\% & 95.0\% & 0.884, 0.454 \\
Apollo & 82.5\% & 83.0\% & 1.569, 0.122 \\
\bottomrule
\end{tabular}
\label{table_global}
\vspace{-8pt}
\end{table}

%% file: discussion.tex
\section{Conclusion}

In this paper, we present a normalizing flow-based framework to solve the robot localization problem. The trained INN provides a bijective mapping between map information and robot poses. While localizing, sampling the latent space gives us a mean and covariance, which can be used as uncertainty estimation for the fusing with other data sources. In our 2D experiments, Local\_INN surpasses particle filer by providing localization with errors as low as 0.032 m and 0.915$^\circ$, while running 270Hz on an embedded platform. Such low latency combined with the fact that its error does not significantly increase with robot velocity makes it suitable for high-speed applications. We also show that Local\_INN has great potential in 3D LiDAR localization with errors of 0.29 m, 0.24$^\circ$ in-session, and 1.41 m, 1.00$^\circ$ out-of-session on the Mulran dataset. Moreover, with our global localization algorithm, Local\_INN has a convergence rate of 93.5\% in the Mulran dataset at the 10th iteration.

%% file: main.bbl
\begin{thebibliography}{10}
\providecommand{\url}[1]{#1}
\csname url@samestyle\endcsname
\providecommand{\newblock}{\relax}
\providecommand{\bibinfo}[2]{#2}
\providecommand{\BIBentrySTDinterwordspacing}{\spaceskip=0pt\relax}
\providecommand{\BIBentryALTinterwordstretchfactor}{4}
\providecommand{\BIBentryALTinterwordspacing}{\spaceskip=\fontdimen2\font plus
\BIBentryALTinterwordstretchfactor\fontdimen3\font minus
  \fontdimen4\font\relax}
\providecommand{\BIBforeignlanguage}[2]{{%
\expandafter\ifx\csname l@#1\endcsname\relax
\typeout{** WARNING: IEEEtran.bst: No hyphenation pattern has been}%
\typeout{** loaded for the language `#1'. Using the pattern for}%
\typeout{** the default language instead.}%
\else
\language=\csname l@#1\endcsname
\fi
#2}}
\providecommand{\BIBdecl}{\relax}
\BIBdecl

\bibitem{dellaert1999monte}
F.~Dellaert, D.~Fox, W.~Burgard, and S.~Thrun, ``Monte carlo localization for
  mobile robots,'' in \emph{Proceedings 1999 IEEE international conference on
  robotics and automation (Cat. No. 99CH36288C)}, vol.~2.\hskip 1em plus 0.5em
  minus 0.4em\relax IEEE, 1999, pp. 1322--1328.

\bibitem{thrun2002probabilistic}
S.~Thrun, ``Probabilistic robotics,'' \emph{Communications of the ACM},
  vol.~45, no.~3, pp. 52--57, 2002.

\bibitem{fox2003bayesian}
V.~Fox, J.~Hightower, L.~Liao, D.~Schulz, and G.~Borriello, ``Bayesian
  filtering for location estimation,'' \emph{IEEE pervasive computing}, vol.~2,
  no.~3, pp. 24--33, 2003.

\bibitem{dube2020segmap}
R.~Dube, A.~Cramariuc, D.~Dugas, H.~Sommer, M.~Dymczyk, J.~Nieto, R.~Siegwart,
  and C.~Cadena, ``Segmap: Segment-based mapping and localization using
  data-driven descriptors,'' \emph{The International Journal of Robotics
  Research}, vol.~39, no. 2-3, pp. 339--355, 2020.

\bibitem{sarlin2019coarse}
P.-E. Sarlin, C.~Cadena, R.~Siegwart, and M.~Dymczyk, ``From coarse to fine:
  Robust hierarchical localization at large scale,'' in \emph{Proceedings of
  the IEEE/CVF Conference on Computer Vision and Pattern Recognition}, 2019,
  pp. 12\,716--12\,725.

\bibitem{tabak2010density}
E.~G. Tabak and E.~Vanden-Eijnden, ``Density estimation by dual ascent of the
  log-likelihood,'' \emph{Communications in Mathematical Sciences}, vol.~8,
  no.~1, pp. 217--233, 2010.

\bibitem{dinh2016density}
L.~Dinh, J.~Sohl-Dickstein, and S.~Bengio, ``Density estimation using real
  nvp,'' \emph{arXiv preprint arXiv:1605.08803}, 2016.

\bibitem{kingma2018glow}
D.~P. Kingma and P.~Dhariwal, ``Glow: Generative flow with invertible 1x1
  convolutions,'' \emph{Advances in neural information processing systems},
  vol.~31, 2018.

\bibitem{papamakarios2021normalizing}
G.~Papamakarios, E.~T. Nalisnick, D.~J. Rezende, S.~Mohamed, and
  B.~Lakshminarayanan, ``Normalizing flows for probabilistic modeling and
  inference.'' \emph{J. Mach. Learn. Res.}, vol.~22, no.~57, pp. 1--64, 2021.

\bibitem{ardizzone2018analyzing}
L.~Ardizzone, J.~Kruse, S.~Wirkert, D.~Rahner, E.~W. Pellegrini, R.~S. Klessen,
  L.~Maier-Hein, C.~Rother, and U.~K{\"o}the, ``Analyzing inverse problems with
  invertible neural networks,'' \emph{arXiv preprint arXiv:1808.04730}, 2018.

\bibitem{ardizzone2019guided}
L.~Ardizzone, C.~L{\"u}th, J.~Kruse, C.~Rother, and U.~K{\"o}the, ``Guided
  image generation with conditional invertible neural networks,'' \emph{arXiv
  preprint arXiv:1907.02392}, 2019.

\bibitem{adler2019uncertainty}
T.~J. Adler, L.~Ardizzone, A.~Vemuri, L.~Ayala, J.~Gr{\"o}hl, T.~Kirchner,
  S.~Wirkert, J.~Kruse, C.~Rother, U.~K{\"o}the \emph{et~al.},
  ``Uncertainty-aware performance assessment of optical imaging modalities with
  invertible neural networks,'' \emph{International journal of computer
  assisted radiology and surgery}, vol.~14, no.~6, pp. 997--1007, 2019.

\bibitem{xiao2020invertible}
M.~Xiao, S.~Zheng, C.~Liu, Y.~Wang, D.~He, G.~Ke, J.~Bian, Z.~Lin, and T.-Y.
  Liu, ``Invertible image rescaling,'' in \emph{European Conference on Computer
  Vision}.\hskip 1em plus 0.5em minus 0.4em\relax Springer, 2020, pp. 126--144.

\bibitem{zhao2021invertible}
R.~Zhao, T.~Liu, J.~Xiao, D.~P. Lun, and K.-M. Lam, ``Invertible image
  decolorization,'' \emph{IEEE Transactions on Image Processing}, vol.~30, pp.
  6081--6095, 2021.

\bibitem{WehRud2021}
T.~Wehrbein, M.~Rudolph, B.~Rosenhahn, and B.~Wandt, ``Probabilistic monocular
  3d human pose estimation with normalizing flows,'' in \emph{International
  Conference on Computer Vision (ICCV)}, Oct. 2021.

\bibitem{kingma2013auto}
D.~P. Kingma and M.~Welling, ``Auto-encoding variational bayes,'' \emph{arXiv
  preprint arXiv:1312.6114}, 2013.

\bibitem{vaswani2017attention}
A.~Vaswani, N.~Shazeer, N.~Parmar, J.~Uszkoreit, L.~Jones, A.~N. Gomez,
  {\L}.~Kaiser, and I.~Polosukhin, ``Attention is all you need,''
  \emph{Advances in neural information processing systems}, vol.~30, 2017.

\bibitem{zhang2018robust}
C.~Zhang, M.~H. Ang, and D.~Rus, ``Robust lidar localization for autonomous
  driving in rain,'' in \emph{2018 IEEE/RSJ International Conference on
  Intelligent Robots and Systems (IROS)}.\hskip 1em plus 0.5em minus
  0.4em\relax IEEE, 2018, pp. 3409--3415.

\bibitem{chen2021icra}
X.~Chen, I.~Vizzo, T.~L{\"a}be, J.~Behley, and C.~Stachniss, ``{Range
  Image-based LiDAR Localization for Autonomous Vehicles},'' in \emph{Proc. of
  the IEEE Intl. Conf. on Robotics \& Automation (ICRA)}, 2021.

\bibitem{chen2020learning}
X.~Chen, T.~L{\"a}be, L.~Nardi, J.~Behley, and C.~Stachniss, ``Learning an
  overlap-based observation model for 3d lidar localization,'' in \emph{2020
  IEEE/RSJ International Conference on Intelligent Robots and Systems
  (IROS)}.\hskip 1em plus 0.5em minus 0.4em\relax IEEE, 2020, pp. 4602--4608.

\bibitem{barsan2020learning}
I.~A. Barsan, S.~Wang, A.~Pokrovsky, and R.~Urtasun, ``Learning to localize
  using a lidar intensity map,'' \emph{arXiv preprint arXiv:2012.10902}, 2020.

\bibitem{lu2019l3}
W.~Lu, Y.~Zhou, G.~Wan, S.~Hou, and S.~Song, ``L3-net: Towards learning based
  lidar localization for autonomous driving,'' in \emph{Proceedings of the
  IEEE/CVF Conference on Computer Vision and Pattern Recognition}, 2019, pp.
  6389--6398.

\bibitem{clark2017vidloc}
R.~Clark, S.~Wang, A.~Markham, N.~Trigoni, and H.~Wen, ``Vidloc: A deep
  spatio-temporal model for 6-dof video-clip relocalization,'' in
  \emph{Proceedings of the IEEE Conference on Computer Vision and Pattern
  Recognition}, 2017, pp. 6856--6864.

\bibitem{uy2018pointnetvlad}
M.~A. Uy and G.~H. Lee, ``Pointnetvlad: Deep point cloud based retrieval for
  large-scale place recognition,'' in \emph{Proceedings of the IEEE conference
  on computer vision and pattern recognition}, 2018, pp. 4470--4479.

\bibitem{cho2022openstreetmap}
Y.~Cho, G.~Kim, S.~Lee, and J.-H. Ryu, ``Openstreetmap-based lidar global
  localization in urban environment without a prior lidar map,'' \emph{IEEE
  Robotics and Automation Letters}, vol.~7, no.~2, pp. 4999--5006, 2022.

\bibitem{sun2020localising}
L.~Sun, D.~Adolfsson, M.~Magnusson, H.~Andreasson, I.~Posner, and T.~Duckett,
  ``Localising faster: Efficient and precise lidar-based robot localisation in
  large-scale environments,'' in \emph{2020 IEEE International Conference on
  Robotics and Automation (ICRA)}.\hskip 1em plus 0.5em minus 0.4em\relax IEEE,
  2020, pp. 4386--4392.

\bibitem{cai2019hybrid}
M.~Cai, C.~Shen, and I.~Reid, ``A hybrid probabilistic model for camera
  relocalization,'' 2019.

\bibitem{kendall2016modelling}
A.~Kendall and R.~Cipolla, ``Modelling uncertainty in deep learning for camera
  relocalization,'' in \emph{2016 IEEE international conference on Robotics and
  Automation (ICRA)}.\hskip 1em plus 0.5em minus 0.4em\relax IEEE, 2016, pp.
  4762--4769.

\bibitem{deng2022deep}
H.~Deng, M.~Bui, N.~Navab, L.~Guibas, S.~Ilic, and T.~Birdal, ``Deep bingham
  networks: Dealing with uncertainty and ambiguity in pose estimation,''
  \emph{International Journal of Computer Vision}, pp. 1--28, 2022.

\bibitem{wei2019learning}
X.~Wei, I.~A. B{\^a}rsan, S.~Wang, J.~Martinez, and R.~Urtasun, ``Learning to
  localize through compressed binary maps,'' in \emph{Proceedings of the
  IEEE/CVF Conference on Computer Vision and Pattern Recognition}, 2019, pp.
  10\,316--10\,324.

\bibitem{mildenhall2021nerf}
B.~Mildenhall, P.~P. Srinivasan, M.~Tancik, J.~T. Barron, R.~Ramamoorthi, and
  R.~Ng, ``Nerf: Representing scenes as neural radiance fields for view
  synthesis,'' \emph{Communications of the ACM}, vol.~65, no.~1, pp. 99--106,
  2021.

\bibitem{sucar2021imap}
E.~Sucar, S.~Liu, J.~Ortiz, and A.~J. Davison, ``imap: Implicit mapping and
  positioning in real-time,'' in \emph{Proceedings of the IEEE/CVF
  International Conference on Computer Vision}, 2021, pp. 6229--6238.

\bibitem{zhu2022nice}
Z.~Zhu, S.~Peng, V.~Larsson, W.~Xu, H.~Bao, Z.~Cui, M.~R. Oswald, and
  M.~Pollefeys, ``Nice-slam: Neural implicit scalable encoding for slam,'' in
  \emph{Proceedings of the IEEE/CVF Conference on Computer Vision and Pattern
  Recognition}, 2022, pp. 12\,786--12\,796.

\bibitem{durkan2019neural}
C.~Durkan, A.~Bekasov, I.~Murray, and G.~Papamakarios, ``Neural spline flows,''
  \emph{Advances in neural information processing systems}, vol.~32, 2019.

\bibitem{wu2020stochastic}
H.~Wu, J.~K{\"o}hler, and F.~No{\'e}, ``Stochastic normalizing flows,''
  \emph{Advances in Neural Information Processing Systems}, vol.~33, pp.
  5933--5944, 2020.

\bibitem{zhang2021diffusion}
Q.~Zhang and Y.~Chen, ``Diffusion normalizing flow,'' \emph{Advances in Neural
  Information Processing Systems}, vol.~34, pp. 16\,280--16\,291, 2021.

\bibitem{winkler2019learning}
C.~Winkler, D.~Worrall, E.~Hoogeboom, and M.~Welling, ``Learning likelihoods
  with conditional normalizing flows,'' \emph{arXiv preprint arXiv:1912.00042},
  2019.

\bibitem{pmlr-v123-o-kelly20a}
M.~O'Kelly, H.~Zheng, D.~Karthik, and R.~Mangharam, ``F1tenth: An open-source
  evaluation environment for continuous control and reinforcement learning,''
  in \emph{Proceedings of the NeurIPS 2019 Competition and Demonstration
  Track}, ser. Proceedings of Machine Learning Research, H.~J. Escalante and
  R.~Hadsell, Eds., vol. 123.\hskip 1em plus 0.5em minus 0.4em\relax PMLR,
  08--14 Dec 2020, pp. 77--89.

\bibitem{walsh17}
\BIBentryALTinterwordspacing
C.~Walsh and S.~Karaman, ``Cddt: Fast approximate 2d ray casting for
  accelerated localization,'' vol. abs/1705.01167, 2017. [Online]. Available:
  \url{http://arxiv.org/abs/1705.01167}
\BIBentrySTDinterwordspacing

\bibitem{betz_autonomous_2022}
J.~Betz, H.~Zheng, A.~Liniger, U.~Rosolia, P.~Karle, M.~Behl, V.~Krovi, and
  R.~Mangharam, ``Autonomous {Vehicles} on the {Edge}: {A} {Survey} on
  {Autonomous} {Vehicle} {Racing},'' \emph{IEEE Open Journal of Intelligent
  Transportation Systems}, vol.~3, pp. 458--488, 2022.

\bibitem{dosovitskiy2017carla}
A.~Dosovitskiy, G.~Ros, F.~Codevilla, A.~Lopez, and V.~Koltun, ``Carla: An open
  urban driving simulator,'' in \emph{Conference on robot learning}.\hskip 1em
  plus 0.5em minus 0.4em\relax PMLR, 2017, pp. 1--16.

\bibitem{kim2020mulran}
G.~Kim, Y.~S. Park, Y.~Cho, J.~Jeong, and A.~Kim, ``Mulran: Multimodal range
  dataset for urban place recognition,'' in \emph{2020 IEEE International
  Conference on Robotics and Automation (ICRA)}.\hskip 1em plus 0.5em minus
  0.4em\relax IEEE, 2020, pp. 6246--6253.

\bibitem{yin2021rall}
H.~Yin, R.~Chen, Y.~Wang, and R.~Xiong, ``Rall: end-to-end radar localization
  on lidar map using differentiable measurement model,'' \emph{IEEE
  Transactions on Intelligent Transportation Systems}, 2021.

\end{thebibliography}
